\documentclass{article} 
\usepackage{iclr2027_conference,times}

\usepackage{amsmath,amsfonts,bm}

\def\eqref#1{equation~\ref{#1}}

\def\1{\bm{1}}

\DeclareMathAlphabet{\mathsfit}{\encodingdefault}{\sfdefault}{m}{sl}
\SetMathAlphabet{\mathsfit}{bold}{\encodingdefault}{\sfdefault}{bx}{n}

\definecolor{pbbg}{HTML}{5B2A9E} 
\definecolor{pbaccent}{HTML}{2A9D8F}

\usepackage{xcolor}
\definecolor{paperteal}{HTML}{328E91} 
\definecolor{jointgcg}{HTML}{A23AA3}
\definecolor{factorizedgcg}{HTML}{2A9D8F}
\definecolor{morphbaseline}{HTML}{1769AA}
\definecolor{linkblue}{HTML}{5292D9}
\usepackage{hyperref}
\hypersetup{
  colorlinks=true,
  linkcolor=pbbg,
  citecolor=pbbg,
  urlcolor=pbbg
}
\usepackage{url}
\usepackage{amsmath,amssymb,graphicx,acro}
\usepackage[nameinlink,capitalise]{cleveref}
\newcommand{\figpanel}[2]{\hyperref[#1]{\cref*{#1}#2}}
\usepackage{placeins}

\acsetup{
  make-links = false,
  first-style = long-short,
}

\DeclareAcronym{gcg}{
  short = GCG,
  long  = Grand Canonical Generators,
}
\DeclareAcronym{gcmc}{
  short = GCMC,
  long  = grand canonical Monte Carlo,
}
\DeclareAcronym{bg}{
  short = BG,
  long  = Boltzmann generator,
}
\DeclareAcronym{snis}{
  short = SNIS,
  long  = self-normalized importance sampling,
}
\DeclareAcronym{gcsnis}{
  short = GC-SNIS,
  long  = grand canonical self-normalized importance sampling,
  short-format = \mbox,
}
\DeclareAcronym{bar}{
  short = BAR,
  long  = Bennett acceptance ratio,
  short-format = \mbox,
}
\DeclareAcronym{cnf}{
  short = CNF,
  long  = continuous normalizing flow,
  short-indefinite = a,
  long-indefinite = a,
}
\DeclareAcronym{cfm}{
  short = CFM,
  long  = conditional flow matching,
}
\DeclareAcronym{dit}{
  short = DiT,
  long  = Diffusion Transformer,
}
\DeclareAcronym{ode}{
  short = ODE,
  long  = ordinary differential equation,
  short-indefinite = an,
  long-indefinite = an,
}
\DeclareAcronym{ess}{
  short = ESS,
  long  = effective sample size,
  short-indefinite = an,
  long-indefinite = an,
}
\DeclareAcronym{mae}{
  short = MAE,
  long  = mean absolute error,
}
\DeclareAcronym{tv}{
  short = TV,
  long  = total variation,
}
\DeclareAcronym{mlp}{
  short = MLP,
  long  = multilayer perceptron,
  short-indefinite = an,
  long-indefinite = a,
}
\DeclareAcronym{ot}{
  short = OT,
  long  = optimal transport,
}
\DeclareAcronym{mic}{
  short = MIC,
  long  = minimum-image convention,
}
\DeclareAcronym{gpu}{
  short = GPU,
  long  = graphics processing unit,
  short-indefinite = a,
  long-indefinite = a,
}
\DeclareAcronym{kl}{
  short = KL,
  long  = Kullback--Leibler,
}
\DeclareAcronym{adaln}{
  short = AdaLN,
  long  = adaptive layer normalization,
  short-indefinite = an,
  long-indefinite = an,
}
\DeclareAcronym{ai}{
  short = AI,
  long  = artificial intelligence,
  short-indefinite = an,
  long-indefinite = an,
}
\DeclareAcronym{lj}{
  short = LJ,
  long  = Lennard--Jones,
}
\DeclareAcronym{lr}{
  short = LR,
  long  = learning rate,
}
\DeclareAcronym{ema}{
  short = EMA,
  long  = exponential moving average,
  short-indefinite = an,
  long-indefinite = an,
}

\title{
\centering
Grand Canonical Generators
}

\newcommand{\authorblock}[2]{{\mdseries\begin{tabular}[t]{@{}c@{}}\textbf{#1} \\ #2\end{tabular}}}
\newlength{\authorcolwidth}
\newcommand{\authorcol}[2]{\makebox[\authorcolwidth][c]{\authorblock{#1}{#2}}}

\author{%
\authorcol{
Andreas Burger
\thanks{This work was conducted during an internship at NVIDIA.}
}{
University of Toronto \\
NVIDIA \\
Vector Institute \\
Toronto, Canada 
}%
\authorcol{Malte Franke}{
ETH Zurich \\
NCCR Catalysis \\
Zurich, Switzerland}%
\authorcol{Luka Mucko}{
University of Toronto \\
Vector Institute \\
Toronto, Canada}%
\\
\rule{0pt}{24pt}\hspace*{0.5\authorcolwidth}%
\authorcol{Kjell Jorner}{
ETH Zurich \\
NCCR Catalysis \\
Zurich, Switzerland}%
\authorcol{Al\'an Aspuru-Guzik}{
University of Toronto \\
NVIDIA \\
Vector Institute \\
Acceleration Consortium \\
CIFAR \\
Toronto, Canada}%
}

\iclrfinalcopy 
\begin{document}

\maketitle
\lhead{} 

\begin{abstract}
We introduce Grand Canonical Generators (GCG), a generative framework that extends Boltzmann generators to the grand canonical ensemble. 
We present two designs. The first conditions a variable-size generative model on the chemical potential, sampling particle number and configuration jointly. The second factorizes the grand canonical distribution into a particle-number distribution and the corresponding canonical Boltzmann density. 
This factorized formulation can use any existing Boltzmann generator for the canonical component, encodes the known linear chemical-potential dependence analytically, and yields a tractable likelihood that supports self-normalized importance sampling (SNIS). 
Empirically, GCG accurately reproduces grand canonical observables on a Lennard--Jones fluid and methane adsorption in a zeolite, demonstrating generalization across chemical potentials and correction via SNIS and grand canonical Monte Carlo.
\end{abstract}

\FloatBarrier
\section{Introduction}\label{sec:introduction}

Sampling molecular systems that exchange particles with an external reservoir is of central interest in problems ranging from gas adsorption in porous materials for carbon capture~\citep{millward2005metal} to solvation free-energy estimation in drug design~\citep{jorgensen1985monte}. 
Such open systems are described by the grand canonical ensemble, whose equilibrium distribution spans both a variable particle number and a continuous molecular configuration at each particle number. 
Accurately computing thermodynamic observables, like adsorption isotherms, solvation free energies, and fluid-phase coexistence, requires sufficient sampling across all relevant particle-numbers.

The dominant approach to sampling open systems is \ac{gcmc}~\citep{adams1975gcmc}, which augments conventional Monte Carlo with particle insertion and deletion moves.
However, trial insertions frequently overlap existing particles, which leads to low acceptance rates and long correlation times~\citep{adams1975gcmc,melling2023enhanced}.
Because Monte Carlo methods move through configuration space in small sequential steps, escaping metastable states and exploring the full space is slow.
The problem worsens when the system spans a wide range of particle numbers: each particle number must be visited often enough to estimate its statistical weight reliably, which further extends simulation times.

\Acp{bg} provide an amortized alternative to molecular simulation, by learning to generate independent, approximately equilibrium configurations in the canonical ensemble~\citep{noe2019boltzmann}.
When the generator provides a tractable likelihood, importance-weighted reweighting corrects residual model error without retraining.
However, existing \acp{bg} fix the particle number and therefore cannot be used for the grand canonical ensemble.

Going from the canonical to the grand canonical ensemble introduces two challenges.
First, the state space becomes a union of configuration spaces of different dimension.
Second, the probability of each particle number depends on the canonical partition functions, which cancel in canonical reweighting and are therefore never estimated by a \ac{bg}.
Because these probabilities shift exponentially with the chemical potential, a grand canonical sampler must implicitly learn free-energy differences across particle numbers.

\looseness=-1
Extending the \ac{bg} paradigm to open systems thus requires 
(1)~a variable-size generative model,
(2)~a particle-number distribution that respects the chemical-potential dependence, 
(3)~a tractable joint likelihood for inference-time correction via importance-sampling. 
No existing method addresses all three requirements.

\textbf{Present work.}
In this work, we introduce \textbf{\acl{gcg}~(\acs{gcg})\acuse{gcg}}, to our knowledge, the first Boltzmann generator for the grand canonical ensemble that operates without lattices and provides a tractable joint likelihood.
We propose two instantiations of that framework that exploit the ensemble structure to different degrees.
\\
A \textbf{Joint GCG} models the complete variable-size state in a single generative process.
It meets requirement (1), but it must learn the chemical-potential dependence from data, and the trans-dimensional Markov process we use provides no tractable likelihood.
\\
A \textbf{Factorized GCG} exploits that at fixed particle number, the grand canonical distribution reduces to the canonical Boltzmann distribution, independent of the chemical potential. It first samples from a learned particle-number model and then from a conditional \ac{bg}.
The particle-number model applies known factors analytically, as in grand canonical histogram reweighting~\citep{ferrenberg1989optimized}, learning only ratios of canonical partition functions, which improves transfer to unseen chemical potentials. This satisfies requirement~(2).
The product of both likelihoods yields a tractable joint likelihood that supports grand canonical \ac{snis}, including a stratified variant that maintains coverage across all relevant particle numbers.
The factorized design thus meets all three requirements.

We validate \ac{gcg} on three systems that probe distinct aspects of the framework.
On an \textbf{ideal gas}, the model recovers the exact particle-number distribution and spatial uniformity, confirming correct variable-size generation without interaction effects.
On a periodic \textbf{Lennard--Jones fluid}, the factorized construction reproduces reference thermodynamic observables after importance-sampling correction and transfers to nearby chemical potentials outside the training range.
On \textbf{methane adsorption in MFI zeolite}, \ac{gcg} generates adsorbed configurations inside the crystalline host and recovers the reference adsorption isotherm.

\begin{figure}[tbp]
  \centering
  \includegraphics[width=0.99\linewidth]{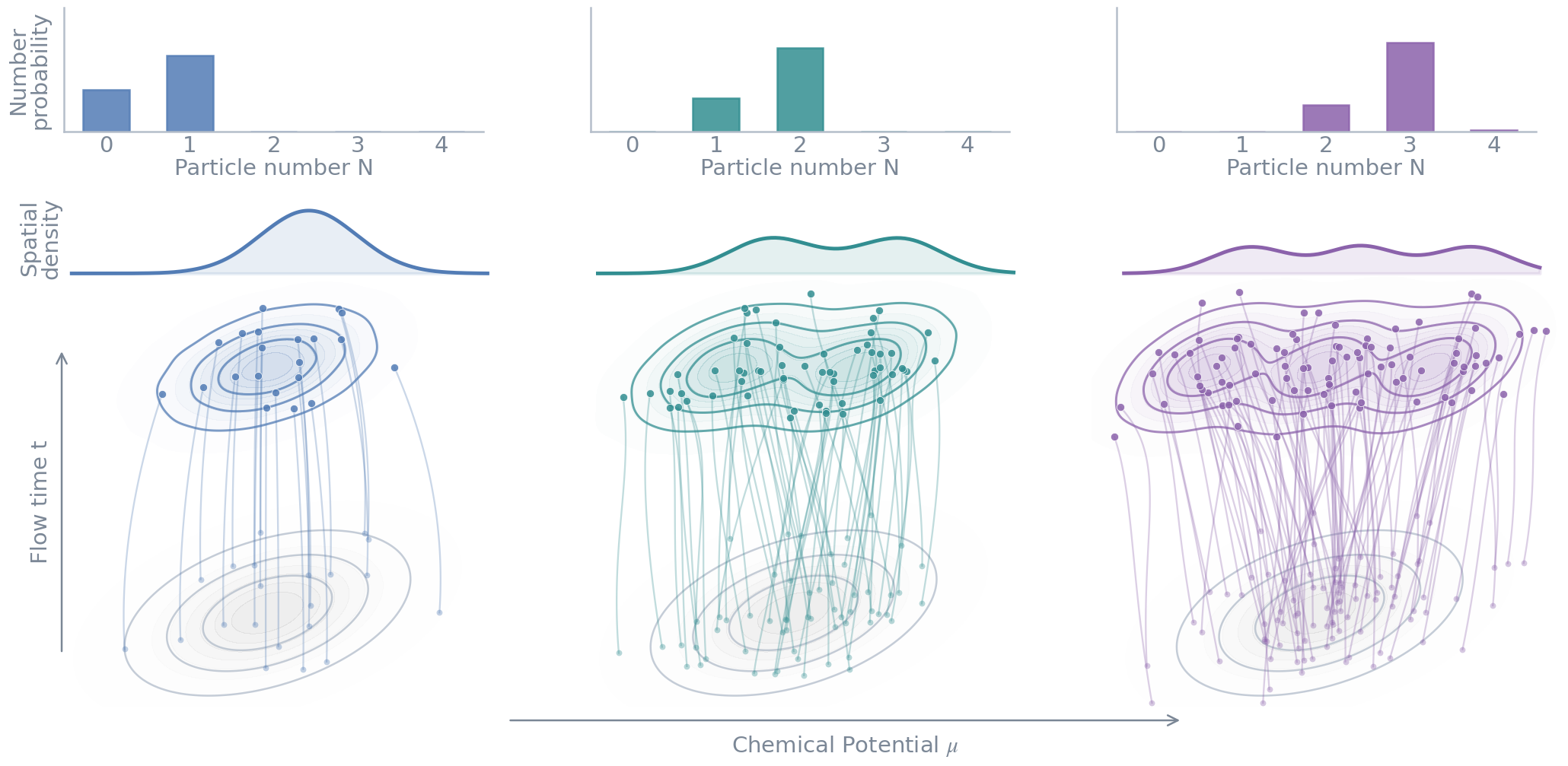}
  \caption{Grand Canonical Generators learn to sample from the $\mu V T$ ensemble, which defines a distribution over both the particle number and geometries, conditioned on the chemical potential $\mu$.
  \vspace*{-0.5cm}
  }
  \label{fig:2d_repulsive_well}
\end{figure}

\FloatBarrier
\section{Background}
We seek to sample from the grand canonical density 
,
also known as the \(\mu V T\)-ensemble
\begin{align}
    &
    \gamma_\mu(N,X;c)
    =
    \frac{\exp[-\beta U_N(X;c)+\beta\mu N]}
       {N!\ \Lambda^{3N}},
\\
    \pi(N,X\mid\beta,\mu,c)
    &=
    \frac{
        \gamma_\mu(N,X;c)
    }{\Xi(\beta,\mu,c)},
\quad
    \Xi(\beta,\mu,c)
=
\sum_{N=0}^{\infty}
\int_{\mathcal V^N}
\gamma_\mu(N,X;c)\,\mathrm{d}X.
    \label{eq:gc_target}
\end{align}
It defines a distribution over both particle numbers \(N\in\mathbb{Z}_{\ge 0}\) and coordinates \(X\in\mathbb{R}^{3N}\), given the inverse temperature \(\beta=1/(k_{\mathrm B}T)\), volume \(V\), and chemical potential \(\mu\). 
The denominator \(\Xi\) is the (intractable) grand canonical partition function.
\(\mathcal V^N\) is the configuration space for \(N\) particles in the volume \(V\), and \(\Lambda\) is the thermal de Broglie wavelength, which accounts for momentum degrees of freedom.
Unlike the canonical ensemble, which fixes \(N\), the grand canonical ensemble allows the system to exchange particles with a reservoir.
The chemical potential \(\mu\) controls this particle exchange and therefore the probability of each particle number.
The context \(c\) describes the surrounding system, e.g. a periodic cell.

\textbf{Boltzmann generators} instead approximate the canonical density at a fixed particle number:
\begin{equation}
\pi(X\mid N,\beta,c)
=
\frac{\exp[-\beta U_N(X;c)]}{Z_N(\beta,c)},
\qquad
Z_N(\beta,c)
=
\int_{\mathcal V^N} \exp[-\beta U_N(X;c)]\,\mathrm{d}X.
\label{eq:canonical_density}
\end{equation}
treating \(N,V,T\) as given and fixed.
The canonical ensemble is also known as the Boltzmann distribution or the \(N V T\)-ensemble.
Typically, a \ac{bg} uses a generative model \(q_\theta(X\mid N,\beta,c)\approx\pi(X\mid N,\beta,c)\)~\citep{noe2019boltzmann} to transform a simple reference distribution into the target distribution through a learned transport map, like a normalizing flow.
Importantly, a \ac{bg} never requires \(Z_N\), since it cancels in self-normalized reweighting.
In contrast, marginalizing \cref{eq:gc_target} over \(X\) yields
\begin{equation}
\pi(N\mid\beta,\mu,c)
=
\frac{e^{\beta\mu N}\, Z_N(\beta,c)}{N!\,\Lambda^{3N}\,\Xi(\beta,\mu,c)},
\label{eq:gc_marginal}
\end{equation}
so the particle-number distribution depends on the intractable \(Z_N\) and shifts exponentially with \(\mu\).
This makes the grand canonical distribution significantly more challenging.

\label{sec:background_flow_matching}
\textbf{Flow matching}~\citep{lipman2022flow} trains \iac{cnf} without backpropagating through its sampling dynamics.
Given source and target samples \(X_0\sim\rho_N\), \(X_1\sim\pi(\cdot\mid N,\beta,c)\), a conditional probability path yields an interpolant \(X_t\) and a target velocity \(u_t(X_t\mid X_0,X_1)\).
A neural velocity field \(v_\theta\) is trained by regressing onto the \ac{cfm} target:
\begin{align}
  \mathcal{L}_{\mathrm{FM}}(\theta)
  =
  \mathbb{E}_{t,X_0,X_1,X_t}
  \bigl\|
    v_\theta(X_t,t;N,\beta,c)
    -
    u_t(X_t\mid X_0,X_1)
  \bigr\|^2.
  \label{eq:background_flow_matching_loss}
\end{align}
At inference, integrating the \ac{ode} \(\dot{X}_t = v_\theta(X_t,t;N,\beta,c)\) from \(t{=}0\) to \(t{=}1\) maps source samples to generated configurations.\label{eq:background_flow_ode}
Under standard regularity conditions, the change-of-variables formula gives the endpoint log-density
\begin{align}
  \log q_\theta(X_1\mid N,\beta,c)
  =
  \log \rho_N(X_0)
  -
  \int_0^1
  \nabla_X\cdot
  v_\theta(X_t,t;N,\beta,c)\,\mathrm{d}t.
  \label{eq:background_cnf_likelihood}
\end{align}
The divergence is the trace of the \(3N\times3N\) velocity-field Jacobian, which dominates the cost of likelihood evaluation and can also be predicted directly~\citep{ouyang2026scallop,burger2025hip}.

\textbf{Self-normalized importance sampling}
\label{sec:background_statistical_correction}
corrects the generative model when the log-density in \cref{eq:background_cnf_likelihood} is available.
Let \((N_i,X_i)\sim q\) for a proposal \(q(N,X\mid\beta,\mu,c)>0\) whenever \(\gamma_\mu(N,X;c)>0\).
\label{eq:background_importance_weight}
For an observable \(\mathcal{O}\), the self-normalized estimator and the associated importance weights are
\begin{equation}
    \widehat{\mathbb{E}}_{\pi}[\mathcal{O}]
    =
    \frac{
    \sum_{i=1}^{K}w_i\,\mathcal{O}(N_i,X_i)
    }{
    \sum_{i=1}^{K}w_i
    }
    ,\qquad
    w_i
=
\frac{\gamma_\mu(N_i,X_i;c)}
     {q(N_i,X_i\mid\beta,\mu,c)}
=
\frac{\exp\!\left[-\beta U_{N_i}(X_i;c)+\beta\mu N_i\right]}
     {N_i!\ \Lambda^{3N_i}\ q(N_i,X_i\mid\beta,\mu,c)}
  .
  \label{eq:background_snis}
\end{equation}
\Ac{snis} is asymptotically unbiased but has finite-sample bias.
The \ac{ess}, \(\operatorname{ESS}=(\sum_i w_i)^2/\sum_i w_i^2\), measures proposal--target overlap.

\textbf{Generator Matching and Morph.}
\label{sec:background_generator_matching}
Flow matching assumes a fixed state dimension.
Generator Matching extends flow matching to time-dependent Markov processes with both continuous and discrete components~\citep{holderrieth2025generator}.
For a variable-size state \(z=(N,X)\), the infinitesimal generator combines coordinate drift with discrete jumps:
\begin{equation}
  \mathcal{A}_t f(z)
  =
  v_t(z)\cdot\nabla_X f(z)
  +
  \int
  \bigl[
    f(z')-f(z)
  \bigr]\,
  K_t(z,\mathrm{d}z'),
  \label{eq:background_markov_generator}
\end{equation}
where \(f\) is an arbitrary test function, \(v_t\) is the continuous drift, and \(K_t\) defines transition rates between states with different particle numbers.
Generator Matching trains the drift and jump components by matching the model to the generator of a conditional probability path.
This loss jointly matches the continuous drift and discrete transition rates, so the learned process follows the desired probability path while allowing changes in particle number.
Morph applies this framework to variable-size molecular generation and uses unbalanced optimal transport (OT) to couple source and target structures of different sizes~\citep{franke2026morphing}.
Unlike \iac{cnf}, the likelihood of a sample using a jump process is generally not tractable, because computing the density requires knowing and integrating over all possible paths that lead to that sample.

\section{Related Work}\label{sec:related_work}
\textbf{Neural equilibrium samplers} learn transformations from simple distributions to independent equilibrium configurations~\citep{noe2019boltzmann,wirnsberger2022normalizing,dibak2022temperature}, with extensions to equivariant and periodic particle models~\citep{klein2023equivariant,hoffmann2026condensed,grenioux2025riemannian}.
When likelihoods are tractable, importance reweighting yields asymptotically correct estimates~\citep{nicoli2020asymptotically,rehman2026autoregressive}.
However, these existing methods assume a fixed particle number.
We extend equilibrium generators to the grand canonical ensemble by learning a tractable particle-number distribution alongside the spatial generator.

\textbf{Learned grand canonical sampling} on lattices has been explored through autoregressive models for semi-grand canonical distributions, where chemical-potential differences change the number of each species at fixed site number~\citep{damewood2022sampling,du2026scaling}, and through Exponentially Tilted Thermodynamic Maps for lattice-gas states across temperature and chemical potential~\citep{lee2026predicting}.
Neural Thermodynamic Integration estimates canonical partition functions across particle numbers but does not provide samples~\citep{mate2024neural}.
JANUS samples site occupations, vacancies, and displacements on crystalline lattice sites, correcting estimates via forward--backward path weights, and learning from energy evaluations instead of precollected samples~\citep{blessing2026janus}.
Grand canonical diffusion changes the number of atoms through continuous voxel fields but targets structure generation~\citep{lei2024grand}.
\citep{christiansen2025gradient} relax discrete atom insertion and deletion into continuous existence variables for gradient-based structure optimization at fixed chemical potential, but do not sample the equilibrium distribution.
Contrary to these works we target continuous particle positions without lattice constraints, and learn a normalized proposal that enables grand canonical importance sampling.

\textbf{Variable-size generation} with permutation-invariant normalizing flows provide tractable densities for unordered point sets if the particle number is supplied~\citep {rasul2019set,bilos2021scalable}.
Autoregressive models generate molecules atom by atom~\citep{cheng2025quetzal,rehman2026autoregressive}, and jump processes insert and delete components during generation~\citep{campbell_trans-dimensional,holderrieth2025generator,franke2026morphing}.
These methods target data generation or molecular design, but have not yet been adapted to the grand canonical ensemble.
As such, they can be used inside the Grand Canonical Generator framework presented in this paper.

\section{Grand Canonical Generators}
\label{sec:gcg}

We introduce \ac{gcg}, a framework for learned sampling in the grand canonical ensemble.
We present two constructions, joint and factorized, that offer different trade-offs between flexibility and likelihood evaluation.
We then describe a structured particle-number model that improves generalization across chemical potentials, and a stratified importance-sampling correction that removes model bias at inference time.
We conclude with the model parameterization used in our experiments.

\textbf{Joint GCG.}
\label{sec:joint_gcg}
The joint construction directly models the complete variable-size state $q_{\psi}^{\mathrm{joint}}(N,X\mid\beta,\mu,c)$. This can be realized using jump diffusion processes, autoregressive models, or, in our case, with the Generator Matching framework from Morph~\citep{holderrieth2025generator,franke2026morphing}.
Its generator changes particle coordinates and particle number within one process, which we condition on $(\beta,\mu,c)$.
Joint \ac{gcg} places few structural restrictions on the learned distribution.
However, it must learn the chemical-potential dependence from data, and its jump process lacks a tractable endpoint likelihood.

\textbf{Factorized GCG.}
\label{sec:factorized_gcg}
Instead we can factorize the grand canonical distribution into particle-number and conditional configuration distributions:
\begin{equation}
  \pi(N,X\mid\beta,\mu,c)
  =
  \pi(N\mid\beta,\mu,c)\;
  \underbrace{
    \frac{\exp[-\beta U_N(X;c)]}{Z_N(\beta,c)}
  }_{\pi(X\mid N,\beta,c)}.
  \label{eq:gc_factorization}
\end{equation}
using the canonical partition function \(Z_N(\beta,c)\) from \cref{eq:canonical_density}.
The conditional configuration distribution \(\pi(X\mid N,\beta,c)\) is now the canonical Boltzmann ensemble, which no longer depends on $\mu$. This allows us to learn a \emph{Factorized \ac{gcg}}:
\begin{equation}
  q_{\phi,\theta}(N,X\mid\beta,\mu,c)
  =
  q_\phi(N\mid\beta,\mu,c)\;
  q_\theta(X\mid N,\beta,c).
  \label{eq:factorized_gcg}
\end{equation}
Sampling first draws $N\sim q_\phi$ and then  $X\sim q_\theta(\cdot\mid N,\beta,c)$.
Conveniently, any \ac{bg} architecture can be conditioned on the particle number and serve as \(q_\theta\).
If \(q_\phi(N\mid\beta,\mu,c)\) and \(q_\theta(X\mid N,\beta,c)\) support likelihood evaluation, the resulting joint likelihood can be used for \ac{snis}.

\textbf{Particle-number model.}
\label{sec:particle_number_model}
Sampling from \(q_{\phi,\theta}(N,X\mid\beta,\mu,c)\) requires a tractable distribution over particle numbers whose likelihood we can evaluate exactly.
A natural choice is a categorical distribution
\(q_\phi(N\mid\beta,\mu,c) = \operatorname{softmax}_N[\ell_\phi(N;\beta,\mu,c)]\)
over \(\{0,\ldots,N_{\max}\}\).
While unrestricted logits suffice in principle, we can exploit the known structure of grand canonical weights to improve transfer across chemical potentials.

At fixed \((\beta,c)\), and with the thermal-wavelength factor \(\Lambda^{3}\) absorbed into the activity \(e^{\beta\mu}\), the grand canonical ensemble assigns each particle number the unnormalized weight
\(\widetilde{p}(N) = e^{\beta\mu N}\,Z_N / N!\)~\citep{adams1975gcmc}.
Taking the logarithm separates two analytically known terms from the unknown canonical partition function \(\log \widetilde{p}(N) = \beta\mu N - \log(N!) + \log Z_N\).
We retain the known terms and represent only the residual \(\log Z_N\) with a learned function \(r_\phi\): \(\ell_\phi(N;\beta,\mu,c) = \beta\mu N - \log(N!) + r_\phi(N,\beta,c)\).
Crucially, \(r_\phi\) does not depend on \(\mu\). Changing the chemical potential applies the exponential tilt \(\beta\mu N\) without retraining.
Transfer to chemical potentials outside the training range thus reduces to extrapolating \(\log Z_N\) to particle numbers outside the training data.
This parameterization mirrors grand canonical histogram reweighting~\citep{ferrenberg1989optimized,wilding1995critical}.
With a free residual per particle number, the maximum-likelihood estimate of \(r_\phi\) on data pooled from several chemical potentials coincides with the multiple-histogram (WHAM/MBAR) estimate of \(\log Z_N\)~\citep{kumar1992weighted,shirts2008statistically}.
The neural residual additionally shares information across \(\beta\) and \(c\), but, as in histogram reweighting, transfer is reliable only where the pooled data cover the relevant particle numbers.

Since the softmax is invariant to additive shifts, the model only needs to learn the ratios $\log(Z_N/Z_0)$.
We fix this gauge freedom by setting \(r_\phi(0,\beta,c)=0\).
We train the particle-number model jointly with \(q_\theta\) by adding the maximum-likelihood loss
to the flow-matching objective.
\begin{align}
  \mathcal{L}_{\mathrm{N}}(\phi)
  &=
  -\mathbb{E}_{(N,\beta,\mu,c)\sim p_{\mathrm{data}}}
  \bigl[\log q_\phi(N\mid\beta,\mu,c)\bigr]
  \label{eq:particle_number_loss}
\end{align}

\textbf{\Ac{gcsnis}.}
\label{sec:gcg_correction}
\looseness=-1
The factorized model provides the joint log-likelihood \(\log q_{\phi,\theta}(N,X)=\log q_\phi(N)+\log q_\theta(X\mid N)\), where we omit the conditioning on \((\beta,\mu,c)\).
Given \(K\) samples \((N_i,X_i)\sim q_{\phi,\theta}\), we can estimate an observable \(\mathcal{O}\) via
\begin{equation}
  w_i
  =
  \frac{\gamma_\mu(N_i,X_i;c)}
       {q_{\phi,\theta}(N_i,X_i\mid\beta,\mu,c)},
  \quad
  \widehat{\mathbb E}_{\pi}[\mathcal{O}]
  =
  \frac{\sum_{i=1}^{K} w_i\,\mathcal{O}(N_i,X_i)}
       {\sum_{i=1}^{K} w_i}.
  \label{eq:gc_snis}
\end{equation}
The grand partition function cancels in the self-normalized ratio.
Direct sampling from \(q_\phi\) can miss particle numbers with small proposal probability but significant target weight.
To address this, we use stratified sampling: we allocate \(n_N\) samples to each particle number \(N\), set \(K=\sum_N n_N\) and \(\alpha_N=n_N/K\), and draw \(X_{N,i}\sim q_\theta(\cdot\mid N,\beta,c)\).
The stratified weight and estimator are
\begin{equation}
  w_{N,i}^{\mathrm{strat}}
  =
  \frac{\gamma_\mu(N,X_{N,i};c)}
       {\alpha_N\, q_\theta(X_{N,i}\mid N,\beta,c)},
  \quad
  \widehat{\mathbb E}_{\pi}^{\mathrm{strat}}[\mathcal{O}]
  =
  \frac{\sum_N\sum_{i=1}^{n_N}
    w_{N,i}^{\mathrm{strat}}\ \mathcal{O}(N,X_{N,i})}
       {\sum_N\sum_{i=1}^{n_N}w_{N,i}^{\mathrm{strat}}}.
  \label{eq:full_stratified_gc_snis}
\end{equation}
\Ac{gcg} samples can also initialize conventional \ac{gcmc} chains.
These chains preserve the target distribution while starting near equilibrium and do not require likelihood evaluations.

\textbf{Architecture.}
\label{sec:model_parameterization}
We implement the joint and factorized \ac{gcg} using Morph~\citep{franke2026morphing} with \iac{dit} backbone.

\textbf{Morph baseline.}
\label{sec:morph_baseline}
We first make minimal changes to adapt Morph to the sampling task and use it as a baseline \citep{franke2026morphing}.
We adapt Morph from molecular design to single-species particle systems by removing bond, atom-type, and charge predictions, as well as bond inputs. We disable center-of-mass removal and condition on $(\beta,\mu,c)$ through the existing \ac{adaln} blocks used for conditioning on the flow time.

\textbf{Joint GCG model.}
\label{sec:joint_gcg_model}
To obtain an improved Joint \ac{gcg}, we make the following changes.
First, we replace the parent-based insertion mechanism with a graph-level insertion intensity and a small \ac{cfm} head for insertion coordinates, which allows the model to handle empty states and spatially homogeneous systems.
The \ac{cfm} head is a small vector-field network attached to the shared \ac{dit} backbone. It transports each insertion position from a uniform periodic source using a short \ac{ode} integration.
Second, we represent coordinates on a periodic torus with a uniform source distribution and wrap coordinates into the simulation cell after each integration step.
The coordinate readout aggregates relative offsets from neighbouring particles within a cutoff using the minimum-image convention.
We choose a cutoff that is smaller than L/2, covered by a smooth envelope, which avoids the discontinuities at the box boundary and at the cutoff, which can hurt model likelihood evaluations.
Third, we construct flow-matching couplings with periodic OT. The assignment cost uses minimum-image distances with Hungarian matching but without Kabsch alignment, which is invalid for a fixed periodic cell.
For the periodic Lennard--Jones experiment, we replace the existing symmetry augmentations with random periodic translations and the 24 cubic rotations.

\textbf{Factorized GCG model.}
\label{sec:factorized_gcg_model}
For Factorized \ac{gcg}, the conditional model \(q_\theta(X\mid N,\beta,c)\) is \iac{cnf} that shares the periodic backbone of Joint \ac{gcg}.
Its velocity network receives \((N,\beta,c,t)\) but not \(\mu\).
The particle-number model \(q_\phi\) uses a small \ac{mlp} for \(r_\phi\) (\cref{sec:particle_number_model}).

For the zeolite adsorption experiment, we continue using the Factorized \ac{gcg} architecture.
Since the zeolite lattice breaks the translational symmetry, we add absolute position features encoded with Fourier features on the torus.
We remove translation and rotation augmentations, replace optimal-transport coupling with independent coupling, and substitute global attention biases with local message passing to improve scalability.

\begin{figure}[hbpt] 
  \centering
  \includegraphics[width=\linewidth]{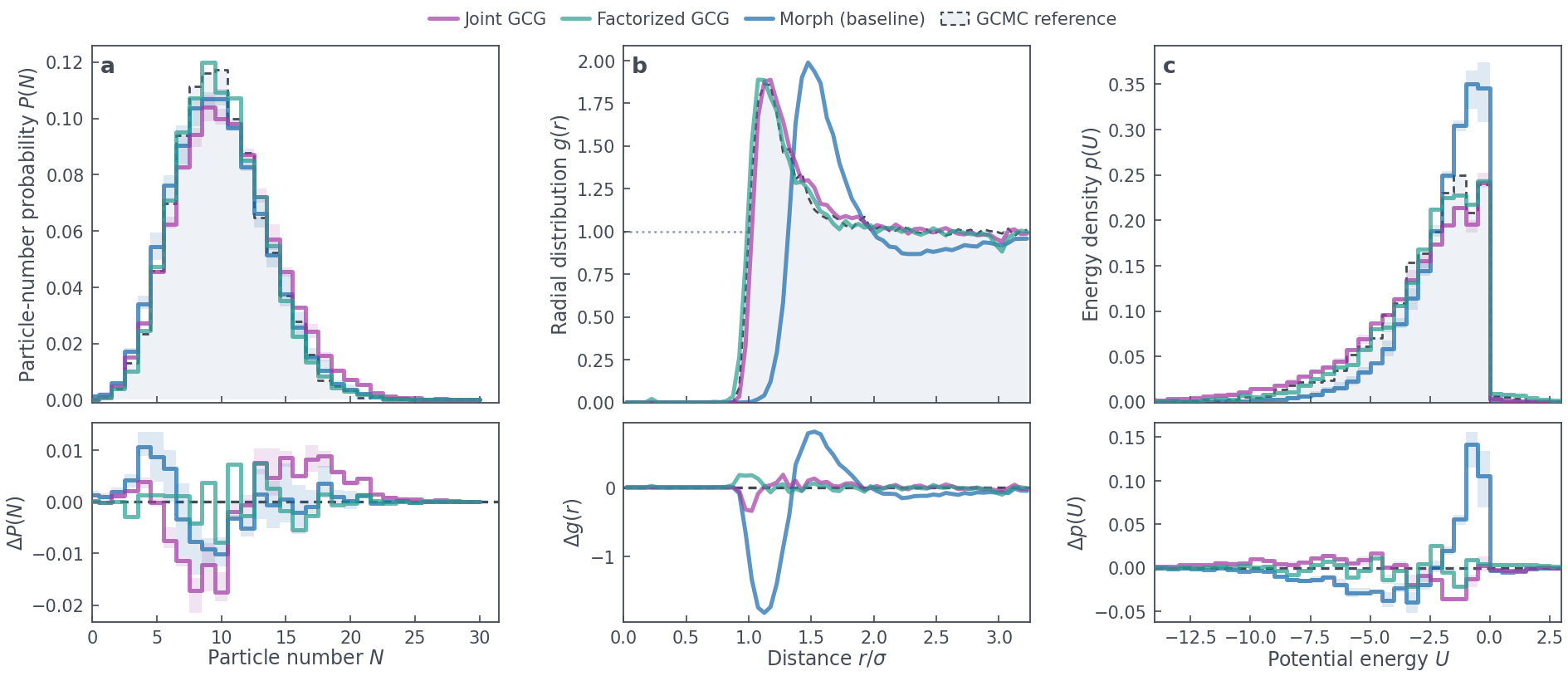}
  \vspace*{-0.6cm}
  \caption{Lennard--Jones.
    Curves and bands show the mean and range over three training seeds.
    }
  \label{fig:lj}
\end{figure}

\section{Experiments}\label{sec:experiments}
Our experiments test whether \ac{gcg} meets the three requirements identified in \cref{sec:introduction}:
(1) Can a variable-size generator sample the joint distribution over $N$ and $X$ of an interacting fluid?
(2) Does the structured particle-number model improve transfer to chemical potentials outside the training range?
(3) Can the tractable likelihood correct residual model error?
Finally, (4) we test whether these properties carry over to a realistic application, methane adsorption in a zeolite.

We compare our \textcolor{jointgcg}{\textbf{Joint \ac{gcg}}} and \textcolor{factorizedgcg}{\textbf{Factorized \ac{gcg}}} against \textcolor{morphbaseline}{\textbf{Morph}}~\citep{franke2026morphing} as a baseline. The models are trained and validated on \ac{gcmc} reference simulations~\citep{hatch2024feasst,dubbeldam2016raspa}.
Datasets and hyperparameters are listed in \cref{app:datasets} and \cref{tab:hyperparameters}.


\begin{figure}
  \centering
  \vspace*{-0.4cm}
  \includegraphics[width=0.8\linewidth]{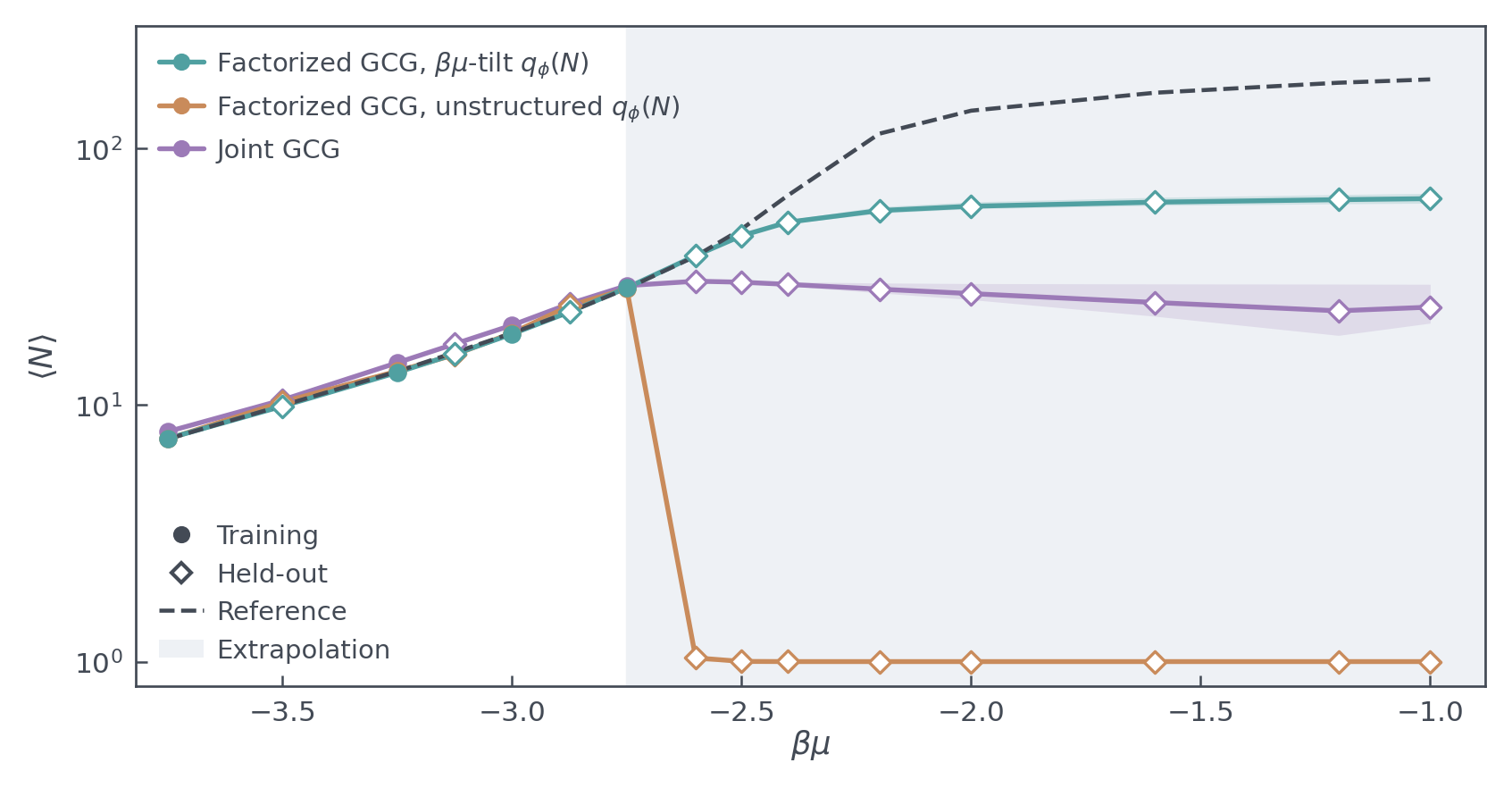}
  \vspace*{-0.6cm}
  \caption{Chemical-potential extrapolation.
  Mean and \(\pm1\) std over three training seeds.
    }
  \vspace*{-0.4cm}
  \label{fig:chemical_potential_transfer}
\end{figure}

\textbf{Lennard--Jones fluid.}
We first ask whether variable-size generators can sample an interacting grand canonical ensemble, and whether an existing variable-size model for molecular design suffices.
We train on a periodic Lennard--Jones fluid at four chemical potentials $\beta\mu\in\{-3.75,-3.25,-3.00,-2.75\}$ and evaluate at the held-out interpolation $\beta\mu=-3.5$.
The units are in terms of the particle diameter $\sigma$ and well depth $\epsilon$ (\cref{app:datasets}).

\looseness=-1
All three models reproduce the particle-number distribution (\figpanel{fig:lj}{a}).
However, Morph places the first peak of $g(r)$ at $r\approx1.5\sigma$ instead of $1.1\sigma$ (\figpanel{fig:lj}{b}) and consequently overestimates the potential energy (mean $-1.9\epsilon$ vs.\ $-2.7\epsilon$, \figpanel{fig:lj}{c}).
Both \ac{gcg} variants recover $g(r)$ and the energy density $p(U)$, with Factorized \ac{gcg} matching the reference most closely and Joint \ac{gcg} slightly overpopulating large $N$ ($\langle N\rangle=10.4$ vs.\ $9.9$).
On an ideal gas without any interactions, both \ac{gcg} variants recover the Poisson particle-number distribution and uniform spatial marginals, whereas Morph overestimates the central density by ${\sim}13\times$ (\cref{app:ideal_gas}).
We attribute Morph's failure to its insertion relative to existing particles and non-periodic coordinates, which bias particle placement.

\textbf{Chemical-potential transfer.}
In applications, a sampler must cover chemical potentials beyond those in its training data, e.g., to extend an adsorption isotherm to higher pressures.
We ask whether building the known $\mu$-dependence into $q_\phi(N)$ enables such transfer to unseen chemical potentials.
To separate the effect of the factorization from the effect of the tilt, we compare Joint \ac{gcg} with two Factorized \ac{gcg} variants: one with an unstructured $q_\phi(N)$ that receives $\mu$ as input, and one with the $\beta\mu$-tilted $q_\phi(N)$ from \cref{sec:particle_number_model}.
We evaluate the learned particle-number distributions on held-out states between the training conditions $\beta\mu\in\{-3.75,-3.25,-3.00,-2.75\}$ and on extrapolation states $\beta\mu\in\{-2.6,-2.5,-2.4\}$, where the reference $\langle N\rangle$ increases from $28.4$ 
to $65.4$.

\looseness=-1
All models reproduce the training states, but differ outside the training range (\cref{fig:chemical_potential_transfer}).
On the held-out states inside the training range, the $\beta\mu$-tilted $q_\phi(N)$ has a relative error of at most $1.1\%$, compared to $8\%$ for Joint \ac{gcg} and $6\%$ for the unstructured $q_\phi(N)$.
The tilted model stays within $0.6\%$ at $\beta\mu=-2.6$ and $6\%$ at $\beta\mu=-2.5$, while Joint \ac{gcg} errs by $21\%$ and $38\%$.
The unstructured $q_\phi(N)$ collapses to $\langle N\rangle\approx1$ at all extrapolation states.
At $\beta\mu=-2.4$, all models fail. The tilted model underestimates $\langle N\rangle$ by $20\%$ and Joint \ac{gcg} by $55\%$.
Thus, the tilt extends the accurate range, but nevertheless accumulates error from extrapolating $\log Z_N$ to unseen particle numbers.
\Cref{app:particle_number_transfer} shows that adding a single-particle reference to $q_\phi(N)$ further improves extrapolation.

\begin{figure}[ht]
  \vspace*{-0.2cm}
  \centering
  \includegraphics[width=0.99\linewidth]{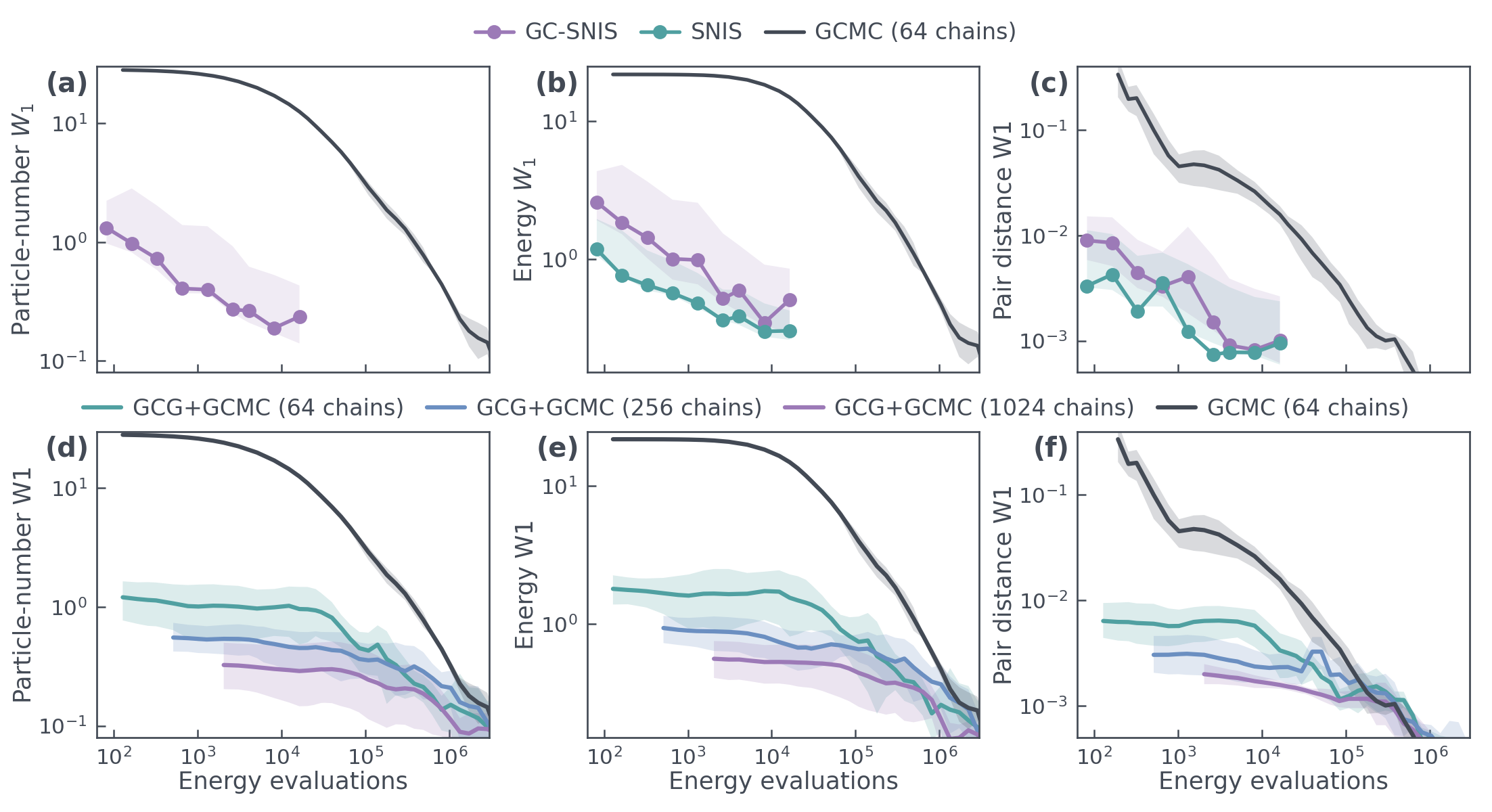}
  \vspace*{-0.4cm}
  \caption{Inference-time correction of Factorized \ac{gcg} on LJ with \ac{snis} (top) and \ac{gcmc} (bottom).
    Bands show $95\%$ bootstrap intervals (200 resamples for \ac{snis}; $2{,}000$ for \ac{gcmc} over five seeds).}
  \label{fig:gc_snis}
\end{figure}

\FloatBarrier
\textbf{Inference-time correction.}
Next, we ask whether the tractable likelihood of a trained Factorized \ac{gcg} can remove residual model bias.
We treat training as a one-time cost that is amortized over chemical potentials and systems in the future, and therefore compare only inference-time estimators.
We draw $400$ configurations for each particle number from $N=10$ to $50$ ($16{,}400$ total), covering $99.7\%$ of the model's particle-number mass.
Conditional \ac{snis} reweights configurations at each fixed $N$, whereas \ac{gcsnis} additionally corrects the particle-number probabilities (\cref{eq:full_stratified_gc_snis}).

\looseness=-1
We measure the error of each scalar observable by its 1-Wasserstein distance $W_1$ to a long \ac{gcmc} reference (\cref{app:metrics}).
With $1.6\times10^4$ target-energy evaluations, conditional \ac{snis} halves the energy $W_1$ ($0.6$ to $0.3$) and lowers the pair-distance $W_1$ from $1.6\times10^{-3}$ to $9\times10^{-4}$.
\Ac{gcmc} needs $92\times$ and $25\times$ more evaluations to match these errors (\cref{fig:gc_snis}, top).
However, \ac{gcsnis} increases the particle-number $W_1$ from $0.08$ to $0.2$.
The learned $q_\phi(N)$ is already accurate on this system and the finite-sample error of \ac{gcsnis} exceeds the model's prediction error.
Both estimators retain $12$--$13\%$ of the configurations as effective samples (\cref{app:ess}).
In \cref{app:bar} we show that estimating the ratios $Z_{N+1}/Z_N$ with the \ac{bar} helps alleviate this error.

\looseness=-1
In order to avoid likelihood evaluations, including Joint \ac{gcg}, generated configurations can instead be used to initialize \ac{gcmc} chains.
We test \ac{gcmc} initialization with the same samples from the Factorized \ac{gcg} model as for \ac{snis}.
Seeded chains start with a $23\times$ lower particle-number $W_1$ than cold-start \ac{gcmc} after two trial moves per chain ($1.2$ vs.\ $28$ for 64 chains).
Their particle-number and energy errors remain below the cold-start baseline until ${\sim}4\times10^6$ evaluations, while pair-distance errors meet at ${\sim}2\times10^5$. 

\textbf{Methane adsorption in a zeolite.}
Finally, we test whether \ac{gcg} carries over to a realistic adsorption problem.
We apply Factorized \ac{gcg} to methane adsorption in a rigid all-silica MFI zeolite, a benchmark system for gas separation.
The crystalline host confines methane to a network of intersecting channels, with up to ${\sim}130$ adsorbed molecules per cell.
We train on seven pressures from $0.1$ to $70\,\mathrm{bar}$ and evaluate at held-out pressures up to $150\,\mathrm{bar}$, twice the largest training pressure. We show the mean and \(\pm1\) std for three training seeds.

\begin{figure}[hbtp]
  \centering
  \includegraphics[width=\linewidth]{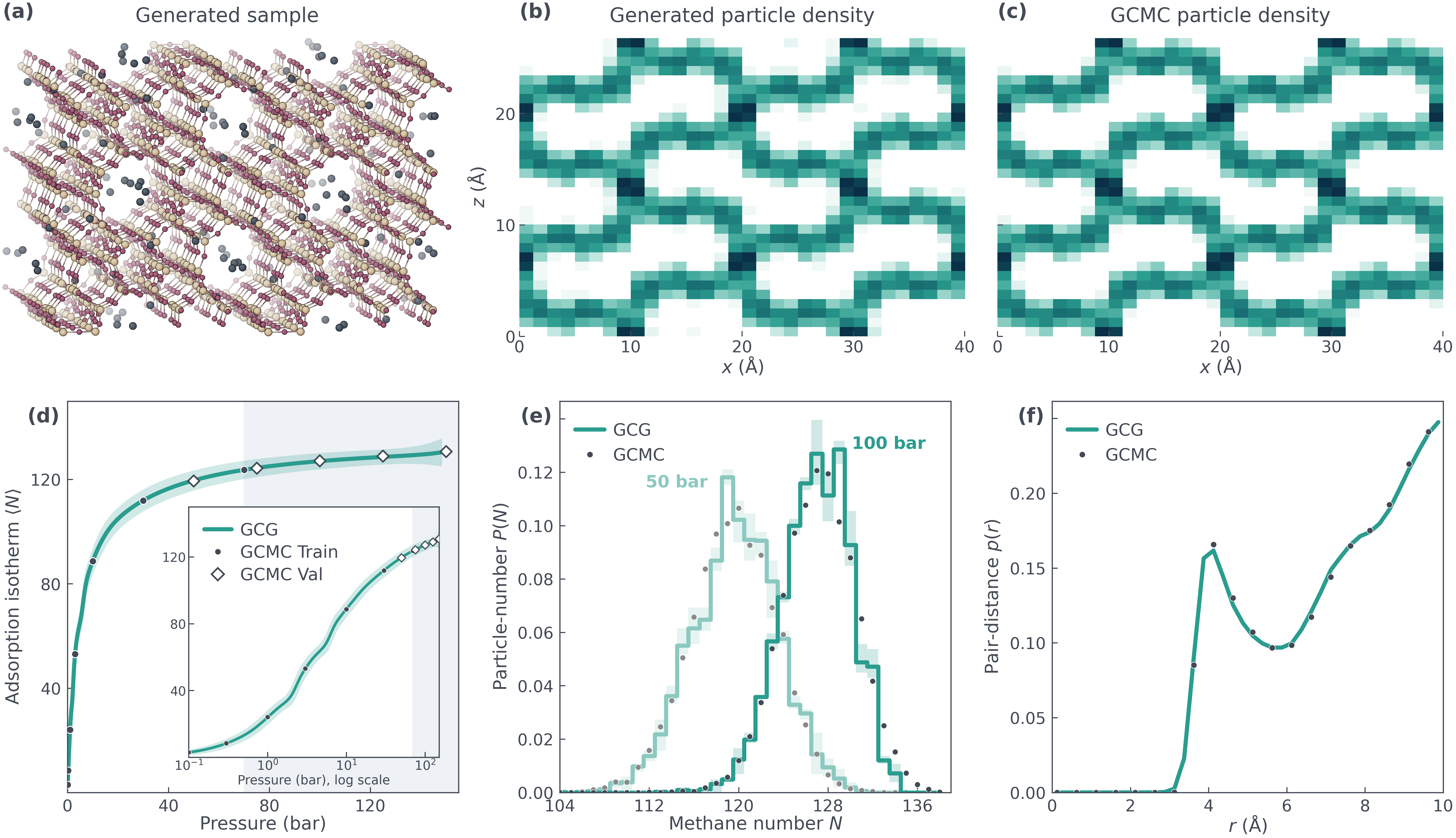}
    \vspace*{-0.6cm}
    \caption{Methane adsorption in MFI zeolite for a Factorized \ac{gcg} model, compared to \ac{gcmc}.
    }
    \vspace*{-0.4cm}
  \label{fig:mfi}
\end{figure}

\looseness=-1
The generated adsorption isotherm agrees with \ac{gcmc} across all held-out pressures, including the extrapolation regime, with the mean error in particle number across three training seeds being at most $0.5$ molecules ($<0.4\%$) (\figpanel{fig:mfi}{d}).
The particle-number distributions also match \ac{gcmc} at $50$ and $100\,\mathrm{bar}$ (\figpanel{fig:mfi}{e}).
At $100\,\mathrm{bar}$, generated configurations reproduce the methane distribution across the channels (\figpanel{fig:mfi}{b,c}) and the methane pair-distance distribution.

\FloatBarrier
\section{Conclusion}
We introduced \acl{gcg}, which extend \acp{bg} from the canonical to the grand canonical ensemble.
Joint \ac{gcg} samples particle number and configuration within a single variable-size process.
Factorized \ac{gcg} imposes additional structure, which leads to favourable results in our experiments.
The Factorized \ac{gcg} combines a particle-number model with any conditional \ac{bg}, enabling a tractable joint likelihood.
By encoding the known chemical-potential dependence analytically, the particle-number model only learns canonical partition function ratios and improves transfer to chemical potentials beyond the training range.
Given a trained model, the tractable likelihood enables grand canonical importance sampling that corrects residual model error, and generated samples accelerate \ac{gcmc} even for models without likelihoods.
On methane adsorption in a zeolite, a Factorized \ac{gcg} accurately reproduces the adsorption isotherm.

Looking forward, the factorized \ac{gcg} is agnostic to the conditional \ac{bg}, so advances in canonical generators carry over.
Autoregressive architectures are a natural fit for the grand canonical ensemble, since they generate variable-size systems particle by particle while providing exact likelihoods, and could unify the joint and factorized variants.
Further directions include simultaneous training on multiple hosts and particles, training on systems with flexible hosts that follow the canonical ensemble, and reducing the reliance on \ac{gcmc} training data through energy-based training objectives.

\textbf{Limitations.}
The \ac{gcg} models presented here require equilibrium samples for training, which have to be collected from \ac{gcmc} simulations.
Amortizing model training across systems remains an open challenge.
We treat methane as a point-like particle and the zeolite crystal as rigid.
While these are standard modelling assumptions in the literature, extending the approach to polar adsorbates such as water, or to flexible frameworks such as proteins, would require sampling over additional degrees of freedom.
Finally, \ac{snis} incurs the cost of evaluating the target log-likelihood for every generated configuration.  Likelihood flow maps could reduce the cost of likelihood evaluation and are a promising direction for future work. Correcting particle-number probabilities with \ac{gcsnis} remains limited by weight variance.

\FloatBarrier

\newpage
\subsubsection*{Acknowledgments}
This work was supported by NVIDIA. 
We would like to thank Austin H. Cheng for fruitful discussions at the start of the project. 
We also thank the  NVIDIA compute infrastructure and cluster operations teams for providing and supporting  the compute resources used for our experiments.
A.A.-G. thanks Anders G. Frøseth for his generous support.
A.B. acknowledges support from the NSERC -Alliance grant ALLRP 587. 
This publication was created as part of NCCR Catalysis (grant number 225147), a National Centre of Competence in Research funded by the Swiss National Science Foundation.
This research was undertaken thanks in part to funding provided to the University of Toronto’s Acceleration Consortium from the Canada First Research Excellence Fund CFREF-2022-00042. 
Resources used in preparing this research were provided, in part, by the Province of Ontario, the Government of Canada through CIFAR, and companies sponsoring the Vector Institute.

\newpage
\bibliography{references}
\bibliographystyle{iclr2027_conference}

\newpage
\appendix
\section{Appendix}
\FloatBarrier

\subsection{Additional results}
\FloatBarrier

\subsubsection{Ideal gas}
\label{app:ideal_gas}

\begin{figure}
  \centering
  \includegraphics[width=\linewidth]{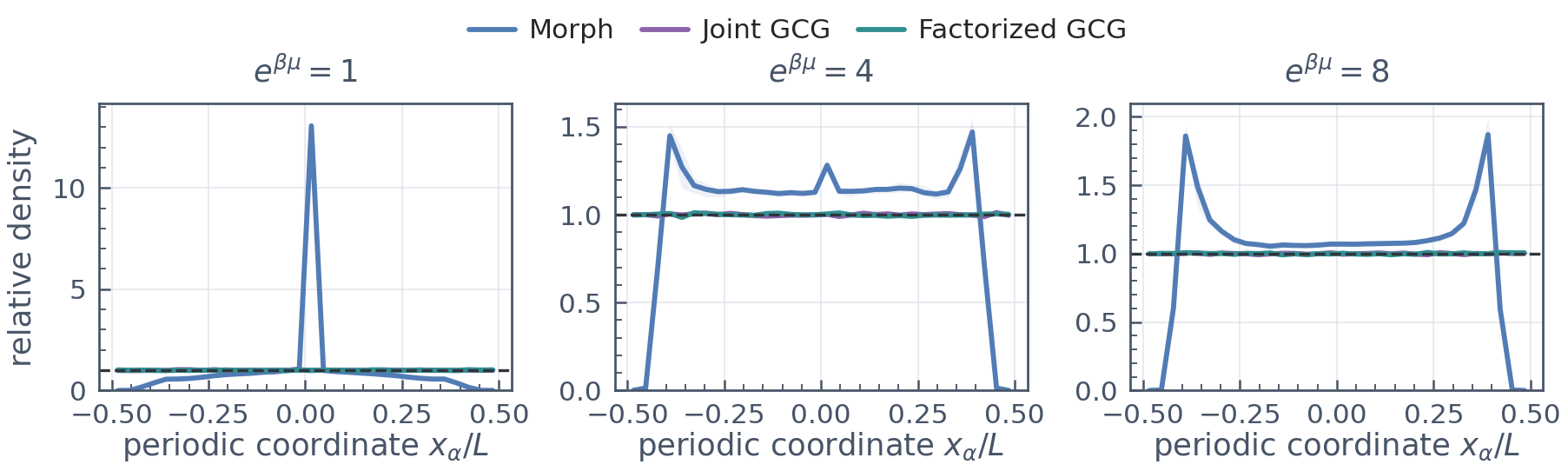}
  \caption{One-coordinate marginals for the periodic ideal gas.
    Lines are the mean density over the three coordinate axes and bands are their range. Dashed is the analytic uniform density.
    Morph develops a sharp central peak at $e^{\beta\mu}=1$.
    Joint \ac{gcg} and Factorized \ac{gcg} remain close to the true marginal at all activities.}
  \label{fig:ideal_gas_coordinate_marginals}
\end{figure}

Previous variable-size generative models were built for molecular design \citep{franke2026morphing}, and may incorporate architectural choices that are less suitable for fluids or gases.
To study any spatial bias separate from interaction effects, we consider the ideal gas, for which the solution is known analytically.

\textbf{Setup.}
For a single-species ideal gas in a periodic box of unit volume ($V=1$, $U_N=0$, $\Lambda=1$), the grand canonical distribution is
\begin{align}
N\sim\operatorname{Poisson}\!\bigl(e^{\beta\mu}\bigr),
  \qquad
  \mathbf{x}_i\mid N\stackrel{\mathrm{iid}}{\sim}\operatorname{Uniform}(\mathcal V),
\end{align}
where \(\mathbf{x}_i\) are the particle positions in \(X=(\mathbf{x}_1,\ldots,\mathbf{x}_N)\) and \(\mathcal V\) is the periodic box. Thus every one-coordinate marginal is uniform and $g(r)=1$.
We train over the continuous range $\beta\mu\in[0,\log 16]$ and evaluate at activities $e^{\beta\mu}\in\{1,4,8\}$ with $65{,}536$ configurations each.

\paragraph{Coordinate total variation.}
We divide each periodic coordinate interval $[-L/2, L/2]$ into $B_x = 32$ equal-width bins. Denoting by $\widehat p_{\alpha b}$ the normalized particle histogram for axis $\alpha \in \{1,2,3\}$ and bin $b$, we report the mean per-axis \ac{tv} distance from the analytic uniform marginal:
\begin{align}
  \operatorname{TV}_{\mathrm{coord}} = \frac{1}{3} \sum_{\alpha=1}^{3} \frac{1}{2} \sum_{b=1}^{B_x} \left| \widehat p_{\alpha b} - \frac{1}{B_x} \right|.
\end{align}

\paragraph{Radial-distribution error.}
We use $B_r = 25$ equal-width radial bins on $0 < r \leq L/2$ with minimum-image pair distances. For radial edges $r_b$, the shell volume is $\Delta V_b = \frac{4\pi}{3}(r_{b+1}^3 - r_b^3)$. If $C_b$ is the observed number of pairs in shell $b$, the ideal-gas normalization, the radial distribution function estimator, and its \ac{mae} are
\begin{align}
  E_b = \sum_i \frac{N_i(N_i-1)}{2} \frac{\Delta V_b}{V}, 
  \qquad 
  \widehat g_b = \frac{C_b}{E_b},
  \qquad
  \operatorname{MAE}_{g} = \frac{1}{B_r} \sum_{b=1}^{B_r} \left| \widehat g_b - 1 \right|
  .
\end{align}
The uniform-sample baseline uses the same $65{,}536$ configurations to measure finite-sample error, which is largest for the low-occupancy inner radial shells.

\textbf{Spatial marginals.}
\Cref{fig:ideal_gas_coordinate_marginals} compares one-coordinate marginals across methods.
Morph develops strong spatial inhomogeneity. At activity one, its central density exceeds the target by a factor of ${\sim}13\times$.
The bias persists at activities four and eight, where Morph places no particles in the outer bins near the box faces and overpopulates the interior.
This is consistent with its non-periodic coordinates and Gaussian source.
Joint \ac{gcg} and Factorized \ac{gcg} both recover nearly uniform marginals at all three activities.

\begin{figure}[htbp]
  \centering
  \includegraphics[width=\linewidth]{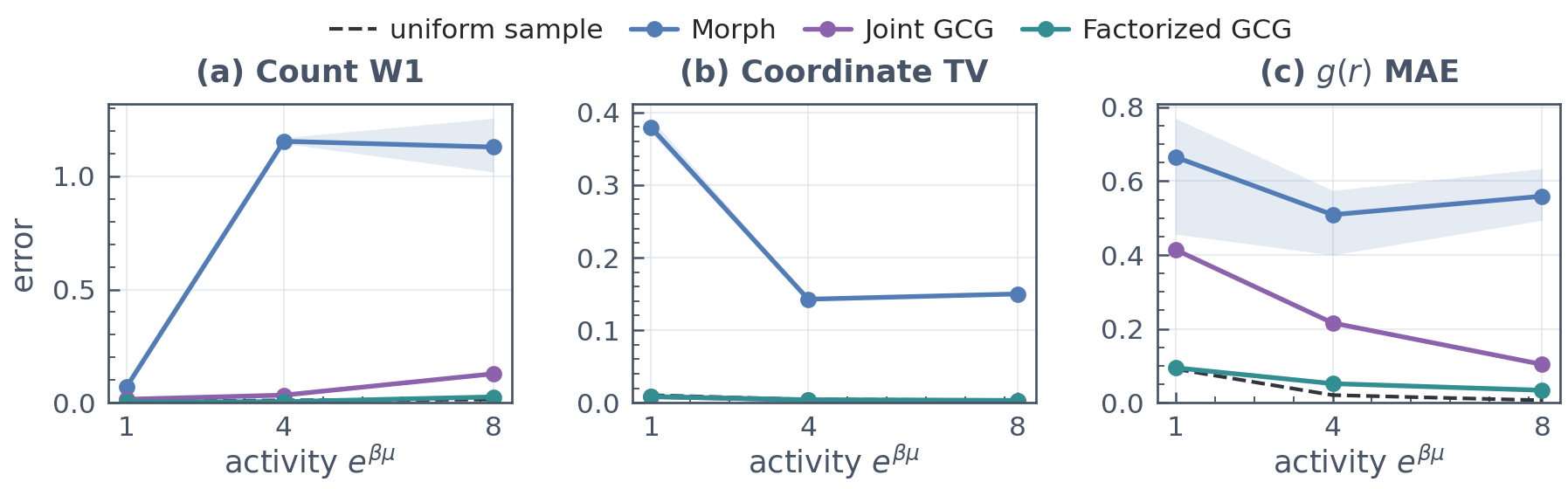}
  \caption{Ideal-gas errors across activity.
    Dashed: finite-sample error from independent uniform configurations.
    }
  \label{fig:ideal_gas_metric_ablation}
\end{figure}

\textbf{Quantitative comparison.}
\Cref{fig:ideal_gas_metric_ablation} reports the one-dimensional Wasserstein distance from the Poisson distribution, mean per-axis coordinate-histogram \ac{tv}, and \ac{mae} of $g(r)$ from one.
At activity four, the particle-number $W_1$ errors are $1.15$, $0.034$, and $0.005$, and the coordinate \ac{tv} errors are $0.143$, $0.0036$, and $0.0041$ for Morph, Joint \ac{gcg}, and Factorized \ac{gcg}, respectively.
Joint \ac{gcg} thus reduces all three errors of Morph, including the $g(r)$ \ac{mae}, from $0.509$ to $0.216$.
However, Joint \ac{gcg} still introduces substantial pair correlations, compared with $0.052$ for Factorized \ac{gcg} and a finite-sample noise of $0.021$.
Factorized \ac{gcg} gives the best overall agreement, while Joint \ac{gcg}'s pair-structure error shows that accurate one-particle marginals alone are not sufficient.

\subsubsection{Particle-number model parameterization}
\label{app:particle_number_transfer}
\begin{figure}[htbp]
  \centering
  \includegraphics[width=0.9\linewidth]{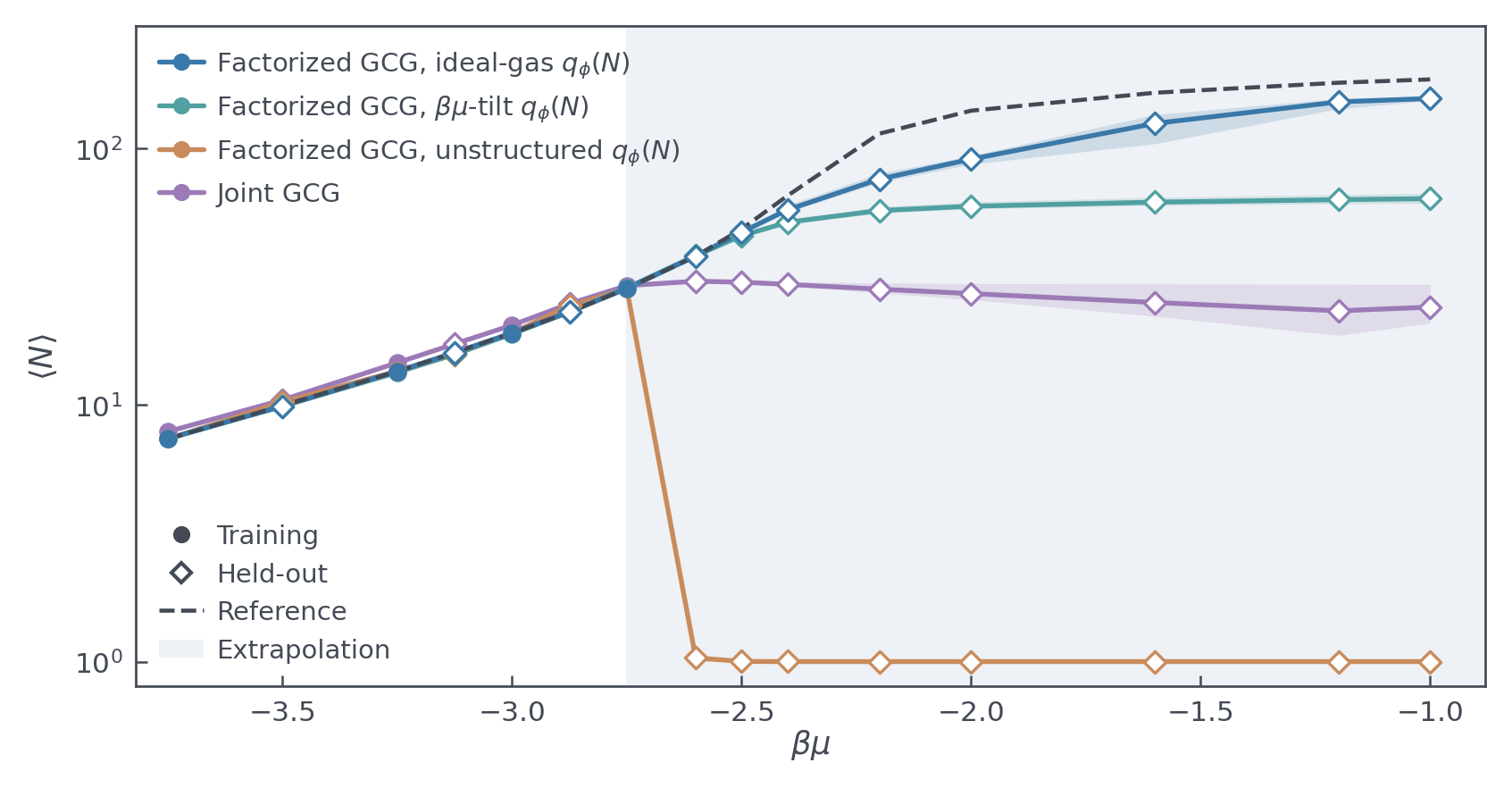}
    \vspace{-0.6cm}
  \caption{Particle-number transfer to unseen chemical potentials.
    \vspace{-1em}
  }
  \label{fig:particle_number_transfer_extended}
\end{figure}

\textbf{Structured particle-number distribution.}
We seek a particle-number model that transfers to unseen chemical potentials and, ideally, to unseen systems.
We start from the $\beta\mu$-tilted model in \cref{sec:particle_number_model}, where $r_\phi$ represents the full $\log Z_N$.
While this applies the exact $\mu$-dependence, $r_\phi$ must also learn the ideal-gas contribution $N\log V$, which grows linearly in $N$.
At initialization, $r_\phi=0$, so the model starts far from any physical distribution and must extrapolate this term to unseen particle numbers.
Fortunately, the ideal gas has the analytic solution $Z_N = q_1^N$, which gives a Poisson particle-number distribution (\cref{app:ideal_gas}).
Writing $\log Z_N = N\log q_1 + \log(Z_N/q_1^N)$ separates this one-particle term from the interaction contribution.
Thus, we generalize the $\beta\mu$-tilted model with an ideal-gas reference:
\begin{align}
  q_\phi(N \mid \beta, \mu, c)
  &= \frac{
    \exp\!\big[
      \beta\mu N + N\log q_1(\beta,c) - \log(N!) + r_\phi(N,\beta,c)
    \big]
  }{
    \sum_{M=0}^{N_{\max}}
    \exp\!\big[
      \beta\mu M + M\log q_1(\beta,c) - \log(M!) + r_\phi(M,\beta,c)
    \big]
  } \notag \\
  &= \operatorname{softmax}_{N}\!\big[
    \beta\mu N + N\log q_1(\beta,c) - \log(N!) + r_\phi(N,\beta,c)
  \big],
  \label{eq:structured_particle_number}
\end{align}
which differs from \cref{sec:particle_number_model} only by the reference term $N\log q_1(\beta,c)$, and reduces to it for $q_1=1$.
Here, $q_1(\beta,c)$ is a one-particle reference partition function, typically in analytic form, that we are free to choose. 
For heterogeneous systems, $q_1(\beta,c)$ represents an accessible volume or a host-specific partition function.
For the homogeneous Lennard--Jones system, we set $q_1 = V$ in reduced coordinates. 

\textbf{Learned residual.}
The residual \(r_\phi\) represents interactions and other deviations from the chosen one-particle reference. 
We parameterize it with \iac{mlp} that receives the particle number and returns one scalar. 
To transfer across systems, an encoder may additionally condition $r_\phi$ on e.g. host and adsorbate features.
As in \cref{sec:particle_number_model}, we absorb \(\Lambda^{3}\) into the activity \(e^{\beta\mu}\).
We fix the additive gauge by enforcing $r_\phi(0,\beta,c)=0$ and initialize the final layer to zero, so the initial distribution coincides with the analytic reference.
Interactions are then learned over the course of training.

\textbf{Results.}
\Cref{fig:particle_number_transfer_extended} extends the main-text figure with a particle-number model that uses the analytic ideal gas as a reference.
Inside the training range, all models except the Joint \ac{gcg} have small errors.
In the extrapolation region, the unstructured model collapses, predicting one $N=1$.
The Joint \ac{gcg} degrades more slowly, with errors from \(20\%\) at \(\beta\mu=-2.6\) to \(55\%\) at \(\beta\mu=-2.4\).
The \(\beta\mu\)-tilted and ideal-gas models stay below \(5\%\) up to \(\beta\mu=-2.5\).
The ideal-gas reference lowers the error further, from \(5\%\) to \(3\%\) at \(\beta\mu=-2.5\) and from \(19\%\) to \(14\%\) at \(\beta\mu=-2.4\), compared to the \(\beta\mu\)-tilted model.
For \(\beta\mu\ge-2.2\), no model is accurate.
The ideal-gas model has the smallest error, at \(13\%\) to \(35\%\).

\subsubsection{Effective sample size}
\label{app:ess}
The \ac{ess} of \ac{snis} depends on the overlap between the proposal and the target.
If the proposal matches the target, all weights are equal and every configuration contributes fully.
If the overlap is small, a few configurations carry most of the weight.
We use the Kish effective sample size~\citep{kish1965survey} of the normalized weights of each estimator,
\begin{align}
  \mathrm{ESS} = \frac{\left(\sum_{i=1}^{M} w_i\right)^2}{\sum_{i=1}^{M} w_i^2},
  \qquad
  w_i =
  \frac{\gamma_\mu(N_i,X_i;c)}
       {q_{\mathrm{prop}}(N_i,X_i\mid\beta,\mu,c)},
\end{align}
which counts how many of the $M$ configurations effectively contribute to the estimate.
Here, \(q_{\mathrm{prop}}\) is the proposal density, including the stratification
factor when applicable.
We report the normalized quantity $\mathrm{ESS}/M$, which is $1$ for a perfect proposal and $1/M$ if a single configuration carries all the weight.
For each sector, $w_i$ are the fixed-$N$ weights of conditional \ac{snis}; for the pooled values, $w_i$ are the weights of conditional \ac{snis} or \ac{gcsnis} (\cref{eq:full_stratified_gc_snis}).

\begin{figure}[tbp]
  \centering
  \includegraphics[width=\linewidth]{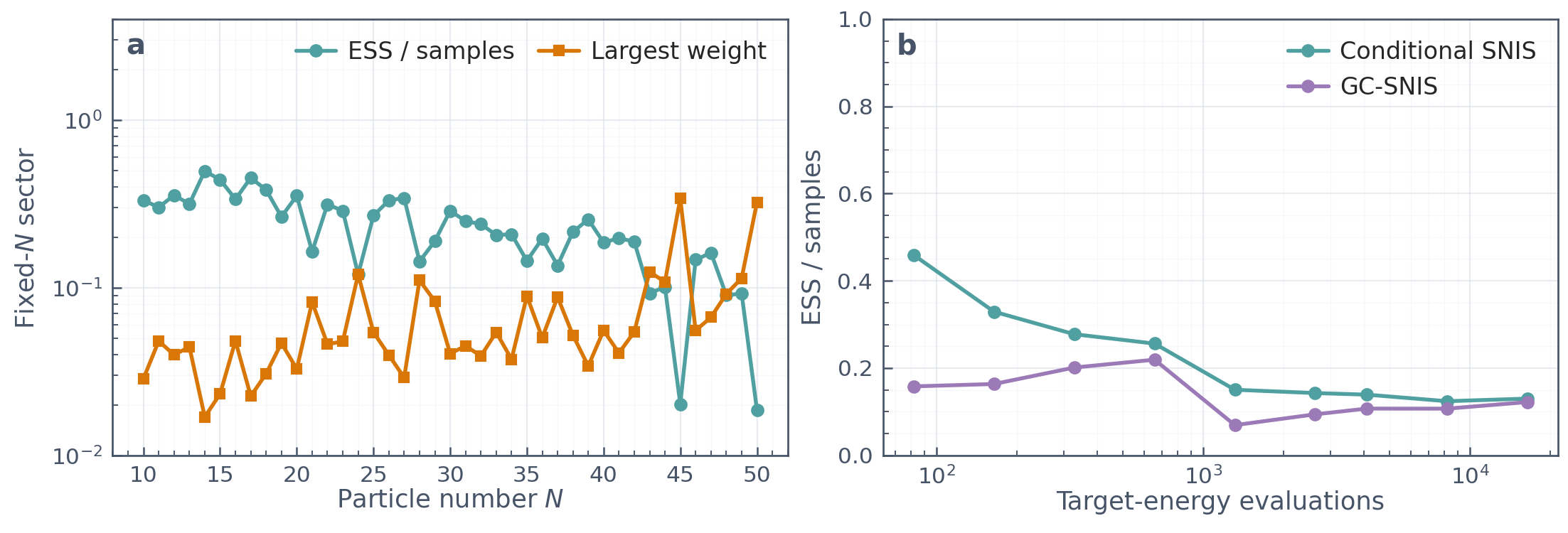}
  \caption{Normalized effective sample size $\mathrm{ESS}/M$ of the stratified pool in \cref{fig:gc_snis}.
    (a)~Per fixed-$N$ sector at $400$ configurations, with the largest normalized weight (dashed).
    (b)~Pooled over all sectors for conditional \ac{snis} and \ac{gcsnis} as a function of target-energy evaluations.}
  \label{fig:ess}
\end{figure}

\Cref{fig:ess} reports this quantity for the stratified pool of \cref{fig:gc_snis}.
Within a sector, $\mathrm{ESS}/M$ ranges from $\sim 0.3$ to $\sim 0.5$ for $N \le 20$ and from $\sim 0.1$ to $\sim 0.2$ for $N \ge 35$.
Only two sectors, $N = 45$ and $N = 50$, fall to $\sim 0.02$, where one configuration carries about one third of the sector weight.
These sectors have model probabilities of only $0.5\%$ and $0.1\%$, so their effect on pooled estimates is small.
Pooled over all sectors, $\mathrm{ESS}/M$ is $0.13$ for conditional \ac{snis} and $0.12$ for \ac{gcsnis} at $16{,}400$ configurations.
The decrease with $N$ is consistent with the growing configurational dimension $3N$.
We leave allocation schemes that adapt to the per-sector \ac{ess} for future work.

\subsubsection{Bennett acceptance ratio}
\label{app:bar}
\textbf{Conditional and grand canonical correction.}
Conditional \ac{snis} reweights configurations at each fixed $N$,
but leaves the particle-number probabilities unchanged.
\ac{gcsnis} uses one global weight and corrects these probabilities, but its
weight variance can be large.

\ac{bar} estimates the ratio \(Z_{N+1}/Z_N\) of canonical partition functions from Boltzmann samples at fixed particle numbers \(N\) and \(N+1\). 
These ratios determine the relative probabilities of observing \(N\) and \(N+1\) particles.
\ac{bar} combines insertion and deletion estimates, which makes the estimate more robust when the two distributions have limited overlap compared to only using insertion or deletion estimates. 
Like for \ac{gcsnis}, we stratify the configuration pool by particle number and allocate a fixed number of configurations to each value of \(N\).

With the canonical partition function $Z_N$ from \cref{eq:canonical_density}, we define
\[
  R_N = \frac{Z_{N+1}}{V Z_N}, \qquad
  \Delta F_N = -\log R_N.
\]
We estimate $\Delta F_N$ from the weighted \ac{bar} equation
\begin{align}
  \sum_i a_i\,s\!\left(\Delta u_i^{+}-\Delta F_N\right)
  =
  \sum_j b_j\,s\!\left(-\Delta u_j^{-}+\Delta F_N\right),
  \qquad
  s(x)=\frac{1}{1+\exp(x)},
\end{align}
where \(\Delta u_i^{+}=\beta[U_{N+1}(X_i\cup\{\mathbf{x}\})-U_N(X_i)]\) is the reduced energy change for inserting a uniform trial particle \(\mathbf{x}\) into a sample \(X_i\) with \(N\) particles, and \(\Delta u_j^{-}=\beta[U_{N+1}(X_j)-U_N(X_j\setminus\{\mathbf{x}_k\})]\) is the change for removing particle \(\mathbf{x}_k\) from a sample \(X_j\) with \(N+1\) particles.
The weights \(a_i\) and \(b_j\) are the \ac{snis} weights of these samples, capped at the 99th percentile for their particle number (\cref{eq:bar_weight_cap}).
The ratio \(R_N\) then gives the relative probability of \(N+1\) and \(N\) particles:
\begin{align}
  \frac{\pi(N+1\mid\beta,\mu,c)}{\pi(N\mid\beta,\mu,c)}
  =
  \frac{\exp(\beta\mu)V}{N+1}\,R_N.
\end{align}
Thus, conditional \ac{snis} corrects the configurations at each fixed $N$, whereas
\ac{bar} corrects the relative probabilities of different $N$.

\begin{figure}[tbp]
  \centering
  \includegraphics[width=\linewidth]{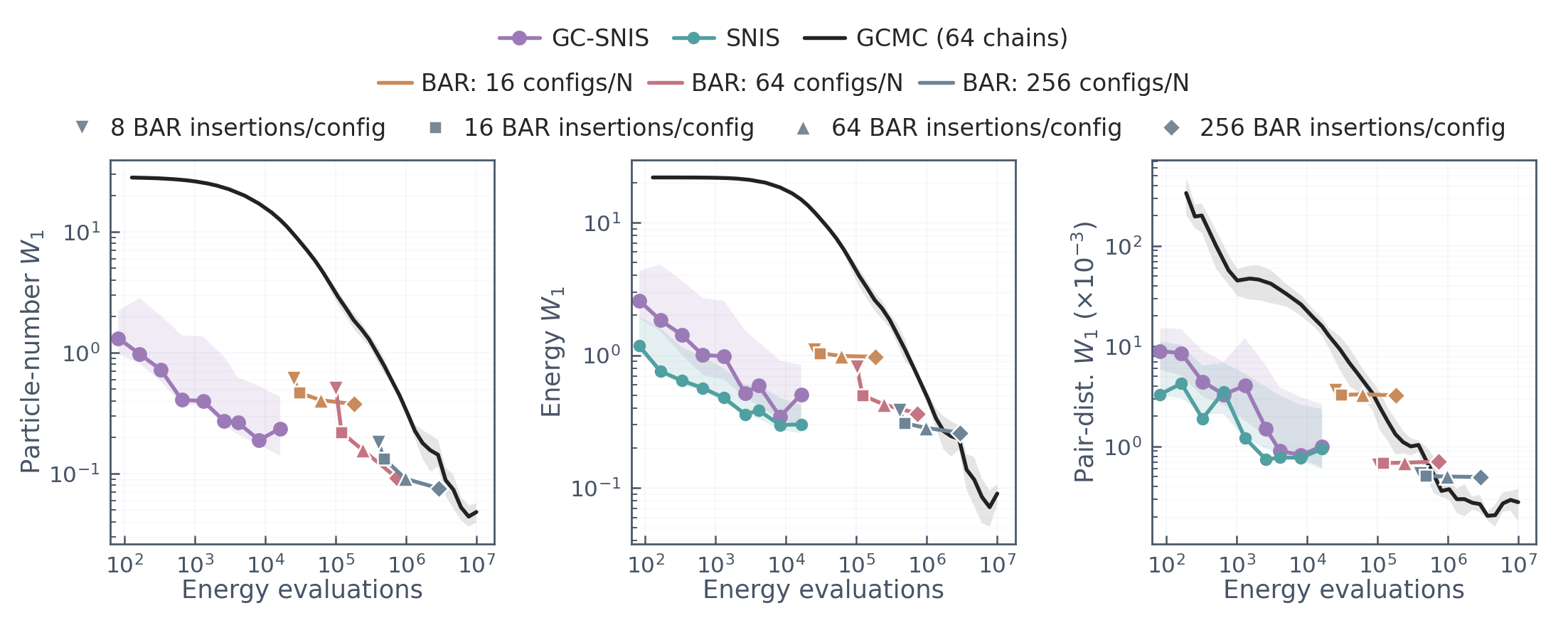}
\caption{Inference-time correction with conditional \ac{snis}, \ac{gcsnis},
    and \ac{bar} between $N$ and $N+1$ particles.
    \Ac{bar} combines insertion and deletion energy changes to reconstruct
    particle-number probabilities.}
  \label{fig:bar}
\end{figure}

\textbf{Weight capping.}
For the \ac{bar} curves, we cap the conditional \ac{snis} weights within each
particle number at their 99th percentile and renormalize them:
\begin{align}
  \widetilde w_i^{(N)}
  = \frac{\min\!\left(w_i^{(N)}, Q_{0.99}(w^{(N)})\right)}
  {\sum_{j: N_j=N}\min\!\left(w_j^{(N)}, Q_{0.99}(w^{(N)})\right)}.
  \label{eq:bar_weight_cap}
\end{align}
We apply the cap before estimating the adjacent particle-number ratios.

\textbf{Results.}
\Cref{fig:bar} compares particle-number, energy, and pair-distance $W_1$
errors against target-energy evaluations.
Lower values indicate better agreement.
Conditional \ac{snis} and \ac{gcsnis} reach low errors through direct
reweighting, whereas \ac{gcmc} improves with the chain budget.
\ac{bar} also improves with more configurations per particle number and
insertion trials, but its performance depends on the proposal budget.

\subsection{Datasets}
\label{app:datasets}
\subsubsection{2D repelling particles in a trap}
For \cref{fig:2d_repulsive_well}, we consider $N$ charged particles at positions $\mathbf{x}_i=(x_i,y_i)$ in an anisotropic harmonic trap, that repel each other via Coulomb interaction
\begin{align}\label{eq:coulomb-trap}
U_N
= \frac{1}{2}\sum_{i=1}^{N}\bigl(x_i^2+16\,y_i^2\bigr)
+ \sum_{i<j}\frac{1}{\sqrt{\|\mathbf{x}_i-\mathbf{x}_j\|^2+\delta^2}},
\end{align}
where the softening length $\delta=0.15$ regularizes the interaction at short range.
The anisotropic confinement is four times stronger along $y$ than along $x$, favoring elongated arrangements.
For a single particle, the trap pulls it toward the center. For two or more particles, the repulsion separates the particles along the $x$ axis.
The plotted flow trajectories represent the generator's transport from its prior.

\subsubsection{Periodic Lennard--Jones fluid}
We consider a single-species Lennard--Jones fluid in a cubic periodic box of side \(L=6.5\sigma\) with particle diameter \(\sigma=1\), well depth \(\epsilon=1\), reduced temperature \(T^*=1.5\) (\(\beta^*=2/3\)), cutoff \(r_c=3\sigma\) (no energy shift, with long-range correction), and thermal wavelength \(\Lambda=1\).

We sample configurations using \ac{gcmc} in FEASST v0.25.19~\citep{hatch2024feasst}, with Metropolis translation, insertion, and deletion moves.
Each trial selects a particle translation or an insertion/deletion move with weights \(1\) and \(2\), respectively.
After \(2{,}000{,}000\) equilibration trials, we record configurations every \(5{,}000\) trials at four training chemical potentials \(\beta\mu\in\{-3.75,-3.25,-3.00,-2.75\}\).

At each chemical potential we generate five independent chains of \(20{,}000\) configurations for training and one chain of \(4{,}096\) configurations for validation, all with independent pseudorandom seeds.
We generate four independent \ac{gcmc} chains at the held-out chemical potential \(\beta\mu=-3.5\), each containing \(8{,}192\) configurations after the same equilibration and recording procedure, for a total of \(32{,}768\) reference configurations.
Coordinates are wrapped into the periodic cell and scaled by \(L/\sqrt{12}\).
We retain particle number, coordinates, total energy, and Lennard--Jones pair energy for every configuration.

\subsubsection{Methane adsorption in MFI zeolite}
We study methane adsorption in a rigid all-silica MFI zeolite.
The framework is the \(2\times2\times2\) supercell of the bundled \texttt{MFI\_SI} structure in RASPA2~\citep{dubbeldam2016raspa}, with box lengths \((40.044, 39.798, 26.766)\,\text{\AA}\).
Methane uses the point-particle treatment from RASPA2 with the bundled \texttt{ExampleZeolitesForceField}.
We use this classical force field for comparability with prior \ac{gcmc} studies. Learned models that provide energies and gradients at higher levels of theory could be used in the future~\citep{eberhard2026derivative,burger2026molelambda}.
All simulations run \ac{gcmc} at \(T = 298\,\text{K}\) with translation, reinsertion, and swap move probabilities of \(0.5\), \(0.5\), and \(1.0\), respectively.

At each pressure, we run \(5{,}000\) initialization cycles followed by \(250{,}000\) production cycles, writing a methane configuration and the particle-number and energy histograms every \(500\) cycles.
We wrap all coordinates into the simulation cell before they enter the models.
We train on \(\{0.1, 0.3, 1, 3, 10, 30, 70\}\,\text{bar}\), and test at 
\(\{50, 75, 100, 125, 150\}\,\text{bar}\). The highest pressure \(150\,\text{bar}\) contains up to \(140\) methane molecules.

\subsubsection{Evaluation metrics}
\label{app:metrics}
For a scalar observable $x$, such as the particle number $N$, the potential energy $U$, or a pair distance $r_{ij}$, we use the 1-Wasserstein distance~\citep{villani2009optimal} between the model distribution $q$ and the \ac{gcmc} reference $p$:
\begin{align}
  W_1(p,q) = \int_{-\infty}^{\infty} \big|F_p(x) - F_q(x)\big|\,\mathrm{d}x,
  \label{eq:w1}
\end{align}
where $F_p$ and $F_q$ are the empirical cumulative distribution functions; for \ac{snis}, $F_q$ uses the normalized importance weights.
$W_1$ measures the average distance that probability mass must move to turn $q$ into $p$.
We report $W_1$ because it has the units of the observable, bounds the error in its mean, $|\langle x\rangle_p-\langle x\rangle_q|\le W_1(p,q)$, and is less prone to outliers than $W_2$.

\subsection{Model and Hyperparameters}
We train each model for 10 to 20 \ac{gpu} hours on an NVIDIA A100.
{\renewcommand{\arraystretch}{1.1}
\begin{table*}[htbp]
  \centering
  \small
  \caption{Model hyperparameters for the \acf{lj} fluid and the MFI zeolite.}
  \label{tab:hyperparameters}
  \resizebox{\linewidth}{!}{%
  \begin{tabular}{lcccc}
    \hline
     & Morph & Joint \acs{gcg} & Factorized \acs{gcg} & Factorized \acs{gcg} \\
    System & \acs{lj} & \acs{lj} & \acs{lj} & Zeolite \\
    \hline
    Temperature
      & \(1.5\) & \(1.5\) & \(1.5\) & 298 K \\
    Box
      & \(6.5^3\) & \(6.5^3\) & \(6.5^3\)
      & \(40.04{\times}39.80{\times}26.77\) \AA \\
    Batch
      & \(1024\) & \(1024\) & \(1024\) & 128 \\
    Steps
      & \(60k\) & \(60k\) & \(60k\) & \(200k\) \\
    Backbone
      & \(8{\times}512\) & \(8{\times}512\) & \(8{\times}512\) & \(8{\times}768\) \\
    Coord. features
      & none & none & none & Torus Fourier \\
    Attention bias
      & yes & yes & yes & no \\
    \(N\) support
      & n.a. & n.a. & \(0-160\) & \(0-160\) \\
    \(N\) model
      & node head & graph intensity & \acs{mlp} \(+\log(N!)\) & \acs{mlp} \(+\log(N!)\) \\
    Optimizer 
      & Muon/AdamW & Muon/AdamW & Muon/AdamW & Muon/AdamW\\
    \acs{lr}
      & \(2\mathrm{e}{-3}/1\mathrm{e}{-2}\)
      & \(2.5\mathrm{e}{-4}/1.2\mathrm{e}{-3}\)
      & \(1\mathrm{e}{-3}/5\mathrm{e}{-3}\)
      & \(7\mathrm{e}{-4}/3.5\mathrm{e}{-3}\) \\
    \acs{ema}
      & \(0.999/5k\)
      & \(0.999/5k\)
      & \(0.999/5k\)
      & disabled \\
    Sample steps 
      & 200 & 200 & 200 & 200 \\
    \hline
  \end{tabular}
  }
\end{table*}
}

\end{document}